\documentclass[journal]{IEEEtran}

\usepackage{amsmath}
\usepackage{subfigure}
\usepackage{caption}
\usepackage{bm}
\usepackage[CJKbookmarks=true]{hyperref}
\usepackage{graphicx} 
\usepackage{epstopdf}
\usepackage{color}
\usepackage{multirow}
\usepackage{amssymb}
\usepackage{times}
\usepackage{epsfig}

\begin{document}
%
\title{M$^5$L: Multi-Modal Multi-Margin Metric Learning for RGBT Tracking}
%
%
%

\author{Zhengzheng Tu, Chun Lin, Chenglong Li\thanks{The authors are with School of Computer Science and Technology, Anhui University, Hefei 230601, China.}, Jin Tang and Bin Luo
}


\markboth{Submission To IEEE TRANSACTIONS ON NEURAL NETWORKS AND LEARNING SYSTEMS}
{Shell \MakeLowercase{\textit{et al.}}: Bare Demo of IEEEtran.cls for IEEE Journals}

\maketitle

\begin{abstract}
Classifying the confusing samples in the course of RGBT tracking is a quite challenging problem, which hasn't got satisfied solution. 
Existing methods only focus on enlarging the boundary between positive and negative samples, 
however, the structured information of samples might be harmed, e.g., confusing positive samples are closer to the anchor than normal positive samples.
To handle this problem, we propose a novel Multi-Modal Multi-Margin Metric Learning framework, named M$^5$L for RGBT tracking in this paper. 
In particular, we design a multi-margin structured loss to distinguish the confusing samples which play a most critical role in tracking performance boosting. 
To alleviate this problem, we additionally enlarge the boundaries between confusing positive samples and normal ones, between confusing negative samples and normal ones with predefined margins, by exploiting the structured information of all samples in each modality.
Moreover, a cross-modality constraint is employed to reduce the difference between modalities and push positive samples closer to the anchor than negative ones from two modalities.
In addition, to achieve quality-aware RGB and thermal feature fusion, we introduce the modality attentions and learn them using a feature fusion module in our network. 
Extensive experiments on large-scale datasets testify that our framework clearly improves the tracking performance and outperforms the state-of-the-art RGBT trackers.
\end{abstract}


%
\IEEEpeerreviewmaketitle

\section{Introduction}
%
%
%
%
\IEEEPARstart{V}{isual} tracking, a fundamental task in computer vision, aims at locating the specific object with a changeable bounding box in the consecutive video frames.
Now visual tracking has applied widely to many fields, such as unmanned aerial vehicle, self-driving cars and video surveillance.
However, there are still many challenges to be solved, specifically when the tracked object receives tremendous influences from the environment. 
RGBT tracking takes advantages of different spectrum data that are visible images and thermal infrared images to allow the object tracked in day and night, and thus receives more and more attentions in recent years~\cite{7577747,LI2019106977,7822984}.
%



\begin{figure}[htbp]
	\centering  
	\includegraphics[width=\linewidth,,height=0.28\textheight]{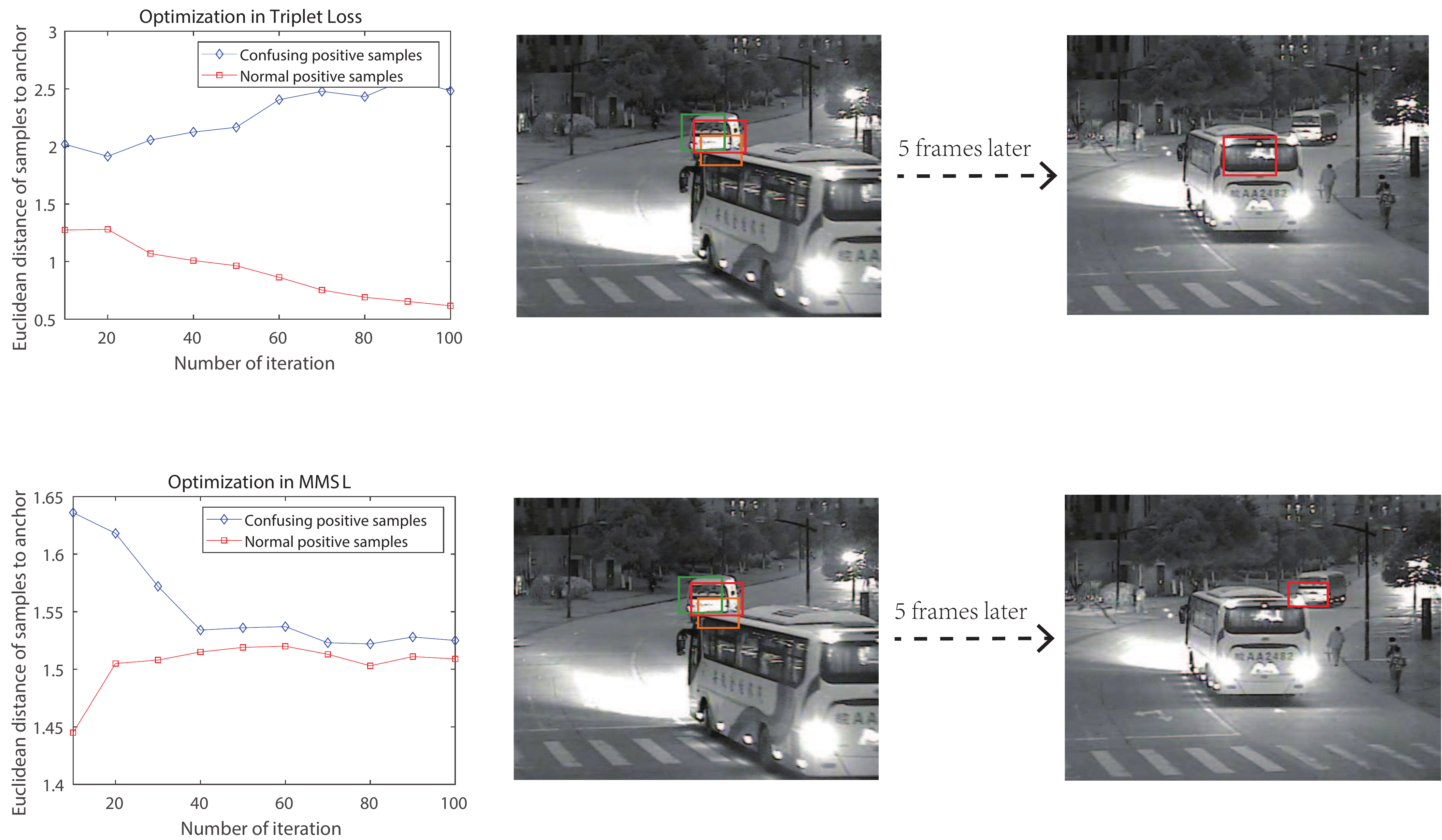}
	\caption{Illustration of Euclidean distance between the anchor and samples.
		With triplet loss, confusing positive samples are not fully optimized, whose distance to anchor might be further than normal negative samples to anchor. While our Multi-modal Multi-margin Structured Loss pushes positive samples closer to anchor and pull negative samples away from anchor, it can preserve structured information of samples with an interval between confusing samples and normal samples.
		Herein, the green, orange and red bounding boxes represent normal sample, confusing sample and tracking result. }  
	\label{structured}  
\end{figure}

Li et al.~\cite{7577747} are the first to propose a comprehensive benchmark datasets to facilitate the research of RGBT tracking. 
Based on the benchmark~\cite{7577747}, they propose a method based on the collaborative sparse representation in the Bayesian filtering framework~\cite{7822984}. 
In the similar tracking framework, Lan et al.~\cite{AAAI1816878} propose to optimize the modality weights for adaptive fusion of different modalities by a max-margin principle on the basis of classification scores.
To improve the robustness of feature representations to background clutter in bounding boxes, Li et al.~\cite{Li:2017:WSR:3123266.3123289} propose a collaborative graph learning algorithm to construct a spatially-ordered weighted patch descriptor and perform object tracking via the structured support vector machine algorithm.
Some works introduce deep learning techniques to boost RGBT tracking performance significantly. 
For example, Li et al.~\cite{long2019multi} present a multi-adapter convolutional neural network to learn the modality-shared, modality-specific and instance-aware target representations.
%
Gao et al.~\cite{Gao_2019_ICCV} adopt an adaptive attention mechanism to fuse useful information from multiple modalities.
However, existing RGBT tracking methods only focus on the fusion of different modalities but ignore the structural information of inter-class and intra-class samples.
Besides, existing RGBT tracking methods also neglect interactive influence between samples from different modalities.

\begin{figure*}[htbp]
	\includegraphics[width=\textwidth,height=0.26\textheight]{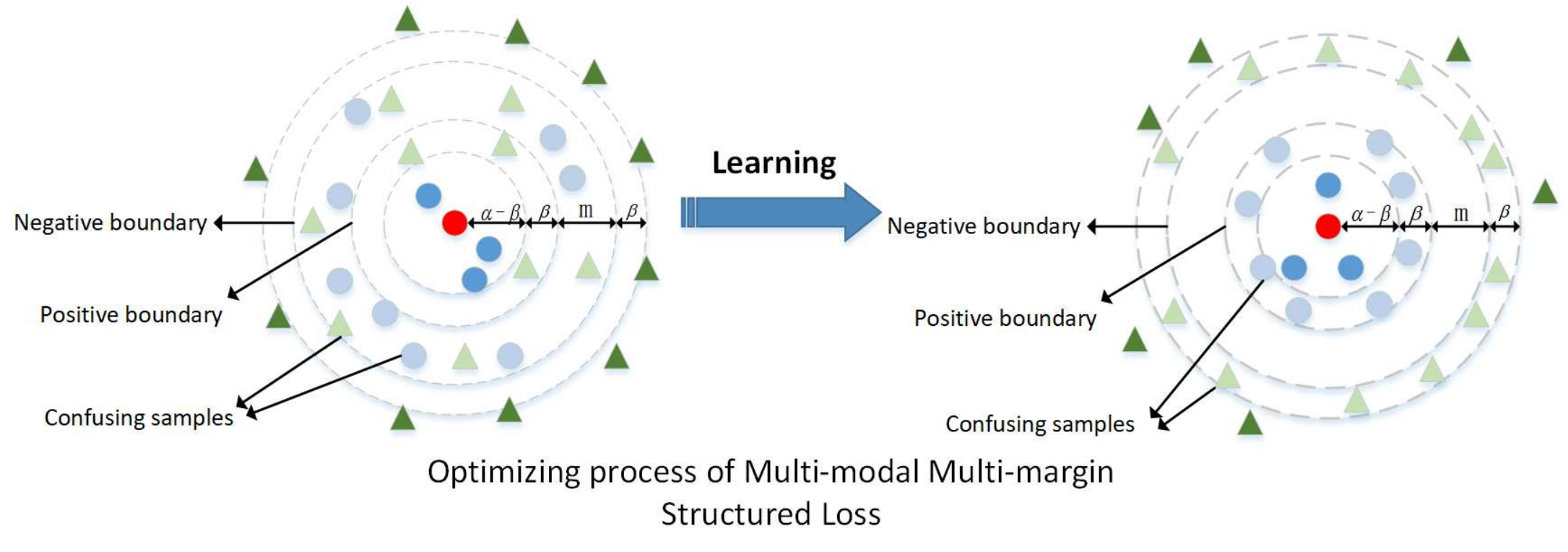}
	\caption{Illustration of optimization process using the Multi-modal Multi-margin Structured Loss.
		The red circle is the anchor (query), and the green triangles represent negative samples while the blue circles represent positive samples.
		It is worth mentioning that the lighter blue circles and lighter green triangles are confusing samples which are hard to discriminate.
		The arrow denotes the gradient direction of optimization.
		The Multi-modal Multi-margin Structured Loss aims to make the confusing positive samples into the interval [$\alpha-\beta,\alpha$] and confusing negative samples into [$\alpha+m,\alpha+m+\beta$] so that positive and negative samples are separated by a margin $m$ and remain original structured information of all samples.}  
	\label{optimise}  
\end{figure*}  

There is no work to address the above problem in RGBT tracking. 
In RGB-based visual tracking, some researches employ the metric learning techniques to maximize the distances between inter-class samples while minimizing the distances between intra-class samples, for performance boosting.
A triplet loss~\cite {Dong_2018_ECCV} is added into Siamese networks by utilizing relationship between positive and negative data.
Hu et al.~\cite{hu2015deep} propose a deep metric learning tracker to learn the nonlinear distance to classify the target and background.
However, these methods only focus on enlarging the boundary between positive and negative samples and have the following issues. 
1) Only a fraction of samples is used in metric learning and some useful samples are thus ignored, making the learned embeddings less discriminative.
2) The structured information of samples(original distribution characteristics of samples) might be harmed, i.e., confusing positive samples are closer to the anchor than normal positive samples, as shown in the first row of Fig.~\ref{structured}.
3) After projecting all samples of multiple modalities into one space, the distances between some positive samples from the same modality to the anchor are farther than some negative ones to the anchor from the other modality, which implies that positive samples are not separated from negative samples after projection.

To alleviate the above problems, in this paper we propose a Multi-Modal Multi-Margin Metric Learning approach (called M$^5$L) for RGBT tracking.
In a specific, we employ all confusing samples to obtain richer structural information of samples.
By optimizing confusing samples in feature space, we make all positive samples closer to the anchor(We regard the ground truth of the target as the anchor, more details can be seen in the sections of framework and experiment) than all negative ones to that, where a margin is set between positive and negative sample sets. 
By this way, the structural information of all samples can be exploited to make the features after embedding more discriminative.
However, simply optimizing all confusing samples might damage the structured information of samples. 
Hence, to preserve structured information of samples as much as possible, we introduce a new loss called Multi-modal Multi-margin Structured Loss.
In the loss, we use multiple margins to enlarge the boundaries between confusing positive samples and normal ones, between confusing negative samples and normal ones with predefined margins, as shown in the second row of Fig.~\ref{structured}.
%
Moreover, with the Multi-modal Multi-margin Structured Loss, the distances between all positive samples to the anchor in each modality will be smaller than the minimum one in the distances between negative samples to the anchor in other modality.
The positive samples can thus be better separated, and thus the classifier will easily find the target object from background.  
The illustration of the optimizing process of our Multi-modal Multi-margin Structured Loss is shown as Fig.~\ref{optimise}.

In addition, we design an adaptive feature integration module to achieve adaptive fusion of different modalities.
In a specific, after extracting features of two modalities, we integrate them according to the modality attentions~\cite{DBLP:journals/corr/abs-1811-09855}, which could make the fusion of different modalities adaptive for more discriminative representations.

\begin{figure*}[htbp]
	\centering  
	\includegraphics[width=\textwidth,height=0.32\textheight]{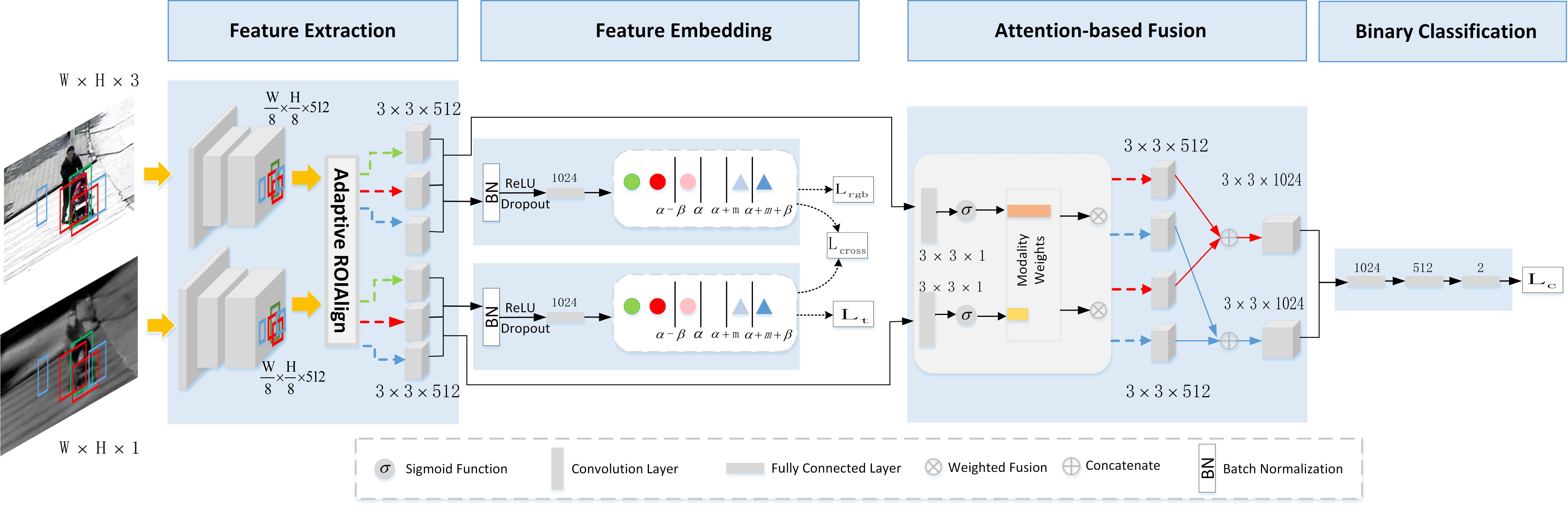}
	\caption{Pipeline of our M$^5$L architecture which contains the feature extraction module, the feature embedding module, the attention-based fusion module and the binary classification module.
		The green, red and blue dotted lines denote the network streams of the anchor, positive and negative samples respectively. 
		The green circle, red circles and blue triangles are corresponding to the anchor, positive samples and negative samples respectively.
		The lighter red circle and lighter blue triangles denote the confusing samples.}  
	\label{network}  
\end{figure*}

The last step of object tracking is to carry out binary classification and bounding box regression~\cite{Nam_2016_CVPR}.
Therefore, the overall loss consists of our multi-modal multi-margin structured loss and the binary classification loss. 
The whole architecture of our network can be seen in Fig.~\ref{network}.

To our best knowledge, it is the first time to investigate the deep metric learning technique for RGBT tracking.
At the same time, our multi-modal multi-margin structured loss implementing the idea of optimizing confusing samples is proved to be effective for improvement of tracking performance.
We summarize the main contributions of this work as follows.
~\\
\begin{itemize}
	\item We propose a deep metric learning based RGBT tracking framework, which exploits the useful information of all samples to boost the performance in RGBT tracking. 
	
	\item We propose a novel Multi-modal Multi-margin Structured Loss to preserve the structured information of samples from RGB and thermal modalities as much as possible and separate all positive and negative samples from all modalities.

	\item We integrate an adaptive feature integration module to achieve adaptive fusion of different modalities in an end-to-end trained deep CNN framework.

	\item Extensive experiments on two large-scale datasets have demonstrated that our RGBT Tracking framework outperforms the state-of-the-art methods.
	
\end{itemize}

\section{Related Work}

RGBT tracking is a challenging topic in the field of computer vision. 
There has been a growing interest in the study of  RGBT tracking in recent years.
In this section, we mainly discuss the following relevant works including RGBT tracking, deep metric learning and deep metric learning based visual tracking.

\subsection{RGBT Tracking}
With the help of thermal infrared modality, visual tracking can track the object better under difficult challenges like occlusion, background clutter and illumination variations.   
Some researches focus on the fusion of RGB and Thermal modalities, and one of the ways is to learn the modality weights to obtain more robust feature fusion.
Li et al.~\cite{7577747,7822984} put forward a method based on the collaborative sparse representation to fuse multi-modal data in the Bayesian filtering framework.
Lan et al.~\cite{AAAI1816878} propose to optimize the modality weights by  max-margin principle on the basis of classification scores.  
However, when classification scores of candidates or reconstructed residues become unreliable, the performance would decrease.

Other researches focus on learning robust representation from a large number of multimodal data~\cite{Li2017Fusing,Li:2017:WSR:3123266.3123289}.
Li et al.~\cite{Li:2017:WSR:3123266.3123289} propose a collaborative graph learning algorithm to construct a spatially-ordered weighted patch descriptor and perform object tracking via the structured support vector machine algorithm.
To fuse the representations from different modalities and avoid introducing noises at the same time, Li et al.~\cite{Li2017Fusing} also create a FusionNet to choose the most discriminative representations from the outputs of two stream ConvNet.   
However, these methods rely on handcrafted features instead of fusing multimodal data with predicting weights from different modalities. 
As the newest researches, Zhu et al.~\cite{acm_zhu} present a recursive strategy(DAPNet) to fuse features of different layers, and Li et al.~\cite{long2019multi} propose a multi-adapter convolutional neural network(MANet) to learn the modality-shared, modality-specific and instance-aware target representations.
Zhu et al.~\cite{Li_2018_ECCV} propose a novel method(CMRT) which depends on a cross-modal manifold ranking algorithm to suppress background effect. 
Zhang et al.~\cite{Zhang_2019_ICCV} propose three novel end-to-end fusion architectures, which consists of pixel-level fusion, response-level fusion, and feature level fusion.  
Unlike these methods, we adopt an adaptive attention-based fusion module~\cite{DBLP:journals/corr/abs-1811-09855} to fuse features from various modalities, meanwhile, multi-modal multi-margin structured loss is proposed to increase classification accuracy.

\subsection{Deep Metric Learning}
Deep metric learning(DML) has been widely used in the tasks of computer vision and also applied to help some challenging classification.In DML, loss function plays a crucial role. Recently, diverse loss functions have been presented~\cite{Schroff_2015_CVPR,NIPS2016_6200,7780803}.Some representive loss functions will be listed as follows.
\par{\bfseries Triplet loss}~\cite{Schroff_2015_CVPR} is made up of three parts:an anchor, a positive sample and a negative sample.The purpose of triplet loss is to pull the positive sample closer to the anchor than the negative and then separate the positive one and negative one by the fixed parameter $m$:
\begin{equation}
\label{s1}
L(X;f)=\frac{1}{\Gamma}\sum\limits_{(i,j,k)\in{\Gamma}}^{\infty}\left\{\begin{array}{ll}
d_{ij}^2+m-d_{ik}^2, &\textrm{if}~~ d_{ij}^2+m-d_{ik}^2 \\
&~~~>0, \\
0 &\textrm{if}~~{otherwise}
\end{array}
\right.
\end{equation}

 where $\Gamma$ is the number of the triplet sets,$i$,$j$ and $k$ are the indexes of the anchor, the poistive point and the negative point respectively.$f$ means embedding function, $d_{ij}$ is the Euclidean metric and $d_{ij}=(f(x_{i})-f(x_{j}))^2$.
\par {\bfseries N-pair-mc}~\cite{NIPS2016_6200} investigates structural characteristics of multiple samples to construct the embedding function.To make a progress based on triplet loss, N-pair-mc pulls one positive sample in positive class from $N-1$ negative samples in $N-1$ negative classes(one negative sample per class):
\begin{align}
\label{s2}
L(&\left\{x_{k},x_{i}\right\}_{i=1}^{N};f)=\nonumber\\
&\frac{1}{N}\sum\limits_{i=1}^{N}\log{(1+\sum\limits_{j\not=i}\exp({f_{k}}^\top{f_{i}}-{f_{k}}^\top{f_{j}} ))}
\end{align}

where $k,i,j$ are indexes of the anchor, the positive sample and negative sample respectively.$\left\{x_{k},x_{i}  \right\}_{i=1}^{N}$ means N sets of samples selected from N different classes.And $f$ is the embedding function.
\par {\bfseries Lifted Struct}~\cite{7780803} unites all negative samples to learn or optimize the embedding function.The main idea of the Lifted Struct is to pull two positive samples chosen randomly as closer as possible and push all negative samples to any of the upper positive pair farther than a parameter(margin) $\beta$:
\begin{align}
\label{s3}
L(X;f)=&\frac{1}{2|P|}\sum\limits_{(i,j)\in{P}}[ d_{ij}+ \log(\sum\limits_{(i,k)\in{N}}\exp(\beta-d_{ik})\nonumber\\
&+\sum\limits_{(j,l)\in{N}}\exp(\beta-d_{jl})]_{+}
\end{align}

\subsection{Deep Metric Learning based Visual Tracking} 
Based on a principled metric learning framework, Wang et al.~\cite{wang2010discriminative} propose a discriminative model taking visual matching and appearance modeling into a single objective, which achieves more persistent results.   
Based on a filter framework, Hu et al.~\cite{hu2015deep} propose a deep metric learning tracker for robust tracking, which learns the nonlinear distance by neural network to classify the target and background.   
In recent research of object tracking, a triplet loss~\cite{Dong_2018_ECCV}  is added into Siamese network by training extracted deep feature maps.   
Compared with traditional pairwise loss, the triplet loss achieves more powerful and discriminative features.
Li et al.~\cite{li2015online} utilize the online distance metric learning to acquire codependent relationship of various feature dimensions.  
Their experiments show that online distance metric learning can improve the robustness of tracking.   
Unlike these methods mentioned above, we present a novel loss in deep metric learning by utilizing the relationship between confusing samples and normal samples to remain structured information of samples for all modalities, and push distance of positive samples to anchor closer than that of negative ones to anchor in RGBT tracking.

\section{M$^5$L Framework}
In this section, we will describe the details of our M$^5$L (Multi-Modal Multi-Margin Metric Learning) tracking framework, including network architecture, training procedure and tracking details.

\subsection{Network Architecture}
The overall network structure of our M$^5$L is illustrated in Fig.~\ref{network}.
M$^5$L mainly consists of four parts, including feature extraction module, feature embedding module, attention-based fusion module and binary classification module.    

{\flushleft \bf Feature extraction module}.
As shown in Fig.~\ref{network}, we use the first three layers of VGG-M network~\cite{Szegedy_2015_CVPR} (denoted as $Conv1-Conv3$) which is pre-trained on the ImageNet dataset.   
Similar to RT-MDNet~\cite{Jung_2018_ECCV}, we discard the max pooling layer in VGG-M to obtain the rich semantic information.
The original RGBT image pair and the bounding boxes around extracted samples are as the input of our network.   
We adopt the adaptive ROIAlign~\cite{He2017Mask} layer in RT-MDNet\cite{Jung_2018_ECCV} to improve the efficiency of feature extraction.     

{\flushleft \bf Feature embedding module}.
To project all features into the same space, we design a feature embedding module, which consists of fully connected (FC) layers and batch normalization (BN).   

To handle the effects of confusing samples, we design a Multi-modal Multi-margin Structured Loss for accurate RGBT tracking.
In the process of the designed loss, we regard the ground truth as anchor, and take positive and negative samples in Gaussian space.   
After inputting all features and the anchor into the feature embedding module, we can easily obtain feature vectors of them. Then, by computing the Euclidean distance between each sample and the anchor, we can easily obtain the similarity between each sample and the anchor.    
We adopt the strategy of mining confusing samples (that are hard to distinguish whether they are positive/negative samples in same feature space) to collect training samples for multi-modal multi-margin structured loss.
Similar to ranked list loss~\cite{DBLP}, we mine confusing samples referred to Eq.~(\ref{s4}) and Eq.~(\ref{s5}):

\begin{equation}
\label{s4}
\begin{array}{l}
{P_{c,i}^{r,*}} = \left\{ x_{c,i}^r| {j}\neq{i},{d_{ij}^r} > \alpha \right\}\\
{N_{c,i}^{r,*}} = \left\{ x_{c,k}^r| {j}\neq{k},{d_{kj}^r} < \alpha + m \right\}
\end{array}
\end{equation}

\begin{equation}
\label{s5}
\begin{array}{l}
{P_{c,i}^{t,*}} = \left\{ x_{c,i}^t| {j}\neq{i},{d_{ij}^t} > \alpha \right\}\\
{N_{c,i}^{t,*}} = \left\{ x_{c,k}^t| {j}\neq{k},{d_{kj}^t} < \alpha + m \right\}
\end{array}
\end{equation}
where $P_{c,i}^{r,*}$ and $N_{c,i}^{r,*}$, $P_{c,i}^{t,*}$ and $N_{c,i}^{t,*}$ are confusing samples in RGB and thermal modalities respectively.$x_{c,i}^r$ and $x_{c,j}^r$ are the feature maps of the anchor and the confusing sample respectively in RGB modality. $d_{ij}^r$ is the Euclidean metric and $d_{ij}^r=(f(x_{c,i}^r)-f(x_{c,j}^r))^2$. $c$ means the confusing sample.  $\alpha$ and $m$ are the positive boundary and the margin between positive and negative boundaries.   

We expect positive samples are close to the anchor as much as possible, while negative samples move against the trail of positive samples.   
Simultaneously, we also expect to guarantee the structured information of samples by means of enlarging boundaries between confusing samples and normal ones.  Mathematically, we formulate the above objective as follows:
\begin{equation}
\label{s6}
\begin{array}{l}
{L}(x_{c,i}^r,x_{c,j}^r;f_r) = \\
~~~~~(1 - {y_{ij}^r})(\left| {\alpha  + m - {d_{ij}^r}} \right| + \left| {\alpha  + m + \beta  - {d_{ij}^r}} \right|)\\
~~~~~ + {y_{ij}^r}(\left| {{d_{ij}^r} - (\alpha  - \beta )} \right| + \left| {{d_{ij}^r} - \alpha } \right|)
\end{array}
\end{equation}
\begin{equation}
\label{s7}
\begin{array}{l}
{L}(x_{c,i}^t,x_{c,j}^t;f_t) = \\
~~~~~(1 - {y_{ij}^t})(\left| {\alpha  + m - {d_{ij}^t}} \right| + \left| {\alpha  + m + \beta  - {d_{ij}^t}} \right|)\\
~~~~~ + {y_{ij}^t}(\left| {{d_{ij}^t} - (\alpha  - \beta )} \right| + \left| {{d_{ij}^t} - \alpha } \right|)
\end{array}
\end{equation}
where $f_{r}$ and $f_{t}$ are the distance metric functions for feature embedding in different modalities. $y_{ij}^r=1$ if $x_{c,i}^r$ and $x_{c,j}^r$ are the features of the anchor and a confusing positive sample respectively in RGB modality.
$y_{ij}^r=0$ if $x_j^r$ is the feature of negative sample.    
From our point of view, positive samples and the anchor belong to the same category, negative samples are not.
$\alpha, \beta, m$ are predefined parameters.
The optimization process is shown in Fig.~\ref{optimise} and the equations are listed as follows:
\begin{equation}
\label{s8}
{L_P^r}(x_{c,i}^r;f_r) = \frac{1}{{\left| {P_{c,i}^{r,*}} \right|}}\sum\limits_{x_{c,p}^r \in P_{c,i}^{r,*}} {{L}(x_{c,i}^r,x_{c,p}^r;f_r)}
\end{equation}
\begin{equation}
\label{s9}
{L_N^r}(x_{c,i}^r;f_r) = \frac{1}{{\left| {N_{c,i}^{r,*}} \right|}}\sum\limits_{x_{c,n}^r \in N_{c,i}^{r,*}} {{L}(x_{c,i}^r,x_{c,n}^r;f_r)}
\end{equation}

\begin{equation}
\label{s10}
{L_P^t}(x_{c,i}^t;f_t) = \frac{1}{{\left| {P_{c,i}^{t,*}} \right|}}\sum\limits_{x_{c,p}^t \in P_{c,i}^{t,*}} {{L}(x_{c,i}^t,x_{c,p}^t;f_t)}
\end{equation}
\begin{equation}
\label{s11}
{L_N^t}(x_{c,i}^t;f_t) = \frac{1}{{\left| {N_{c,i}^{t,*}} \right|}}\sum\limits_{x_{c,n}^t \in N_{c,i}^{t,*}} {{L}(x_{c,i}^t,x_{c,n}^t;f_t)}
\end{equation}
where $P_{c,i}^{r,*}$ and $N_{c,i}^{r,*}$ are confusing samples in RGB modality which are decided by Eq.~(\ref{s4}) in all samples.  
$x_{c,p}^r$($x_{c,p}^t$) and $x_{c,n}^r$($x_{c,n}^t$) belong to the sets of confusing positive and negative sample respectively in RGB(T) modality.
%

Enlarging the distance between positive and negative samples in each modality is not sufficient, since distances between some positive samples and the anchor may be farther than distances between some negative samples and the anchor when they are in different modalities. 
Therefore, we further propose the cross-modality constraint between two modalities, which is based on the triplet loss~\cite{Schroff_2015_CVPR};
\begin{equation}
\label{s12}
\begin{array}{l}
{L_{cross}} =\max \left\{d^r(P_r^+)-\min \left\{d^t(N_t^-) \right\}+\delta,0\right\} \\
~~~~~~~~~~~+\max \left\{d^t(P_t^+)-\min \left\{d^r(N_r^-) \right\}+\delta,0\right\}
\end{array}
\end{equation}
where $d^r(P_r^+)$ denotes the Euclidean distance between the positive sample and the anchor in RGB modality.
Likewise, $d^t(N_t^-)$ is the Euclidean distance between the negative sample and the anchor in thermal modality.
$\delta$ is an enforced margin between a pair of positive and negative samples. 

In Multi-modal Multi-margin Structured Loss (MMSL), we minimize objectives($ {L_{rgb}}(x_{c,i}^r;f_r)$, ${L_t}(x_{c,i}^t;f_t)$, $L_{cross}$) equally and jointly optimize them:
\begin{equation}
\label{s13}
{L_{rgb}}(x_{c,i}^r;f_r) = {L^r_P}(x_{c,i}^r;f_r) + {L^r_N}(x_{c,i}^r;f_r)
\end{equation}
\begin{equation}
\label{s14}
{L_{t}}(x_{c,i}^t;f_t) = {L^t_P}(x_{c,i}^t;f_t) + {L^t_N}(x_{c,i}^t;f_t)
\end{equation}
\begin{equation}
\label{s15}
{L_{MMSL}} = {L_{rgb}}(x_{c,i}^r;f_r) + {L_t}(x_{c,i}^t;f_t) + L_{cross}
\end{equation}
where MMSL belongs to a part of the total loss, which plays a crucial role in training.

{\flushleft \bf Attention-based fusion module}.
Due to different properties and changing quality of RGB and thermal modalities, 
several methods~\cite{7577747,7822984,AAAI1816878} propose to compute weights for different modalities.   
However, these methods may misclassify the object because of bad modality weight or unreliable computations.
In this paper, we adopt an attention-based fusion scheme like~\cite{DBLP:journals/corr/abs-1811-09855} to compute importance of each modality in an end-to-end CNN framework.   
To explicitly measure the importance of features in different modalities, we introduce modality attentions to fuse them adaptively, and all attentions are collaboratively learned, similar to adaptive feature recalibration in~SENet\cite{hu2018squeeze}.    
After obtaining the modality attentions, we aggregate feature maps of different modalities adaptively according to their attentions.     
Specifically, this module includes a convolution layer, a ReLU activation function and a sigmoid function.    
More details are presented in Fig.~\ref{network}.
We briefly use the following formulas to deliver the process of attention-based fusion module:
\begin{equation}
\label{s16}
{x^{cat}} = (\sigma ({D_R}) \otimes {D_R}) \oplus (\sigma ({D_T}) \otimes {D_T})
\end{equation}
where $D_R$ and $D_T$ are the features of RGB and thermal modalities from feature extraction module, $\sigma(\cdot)$ is sigmoid function, $\otimes$ and $\oplus$ are weighting operation and concatenate operation respectively, and $x^{cat}$ is the concatenated RGBT feature.

{\flushleft \bf Binary classification module}.
In this module, we use three fully connected layers with ReLUs and dropouts for binary classification.  
Softmax cross-entropy loss is adopted as binary classification and we select the higher classification scores as candidate bounding boxes of tracking.   
More training and tracking details can be found in next section.

\subsection{Training Procedure}
The whole network is end-to-end trained.   
We first take the image, the ground-truth bounding box, 64 positive proposals and 196 negative proposals extracted from the Gaussian space as the inputs of the network.   
To be specific, 64 positive and 196 negative proposals are drawn according to the overlap ratios($IoUs$) with the ground-truth bounding box, where bounding boxes with $IoUs$ larger than 0.7 are treated as positive samples and $IoUs$ of negative samples are less than 0.5.   
Then, we use two pre-trained VGG-M models to extract the first three layer features of RGB and thermal images respectively, and adopt adaptive ROIAlign~\cite{He2017Mask} mapping the ground truth and proposals to the feature maps of RGB and thermal modalities.   
After entering feature embedding module, we can obtain the multi-modal multi-margin structured loss.  
At the same time, fused features trained through attention-based fusion module are used to obtain the binary classification loss according to three fully connected layers.   
Ultimately, the Multi-modal Multi-margin Structured Loss and binary classification loss are combined together to jointly optimize the whole network.

In the feature embedding module, we set $\alpha$ and $\beta$ to 1.6 and 0.1 respectively.   
And the margin($m$) between positive and negative examples are set to 0.2.
These parameters are validated through experiments.   
With respect to network training, the learning rate is 0.0001 for convolutional layers and 0.001 for fully connected layers.   
We alternatively train and test our M$^5$L on the dataset \textbf{GTOT}~\cite{7577747} and \textbf{RGBT234}~\cite{LI2019106977}.

\subsection{Tracking Details}
In the course of tracking, major processes are consistent with training phase. 
Attention-based fusion module is used to integrate RGBT feature maps after extracting features of different modalities.   
Then these feature maps are sent to three fully connected layers, and it is worth mentioning the last fully connected layer is responsible for each test sequence.   
As same as MDNet does, we also adopt the principles of short-term and long-term updates~\cite{Nam_2016_CVPR}.   
Given the $j$-th frame, we update 256 candidates from a Gaussian distribution of last tracking result.   
For each candidate, we compute its positive and negative scores, and the target location in current frame is defined by the candidate with the maximum positive score.   
Eventually, we also adopt the technique of bounding box regression~\cite{Nam_2016_CVPR} to improve the location accuracy.

\subsection{Discussion}
Here, we will discuss the difference between ranked list loss~\cite{DBLP} and our Multi-modal Multi-margin Structured Loss.In ranked list loss, instead of a certain positive data point,the whole positive sample set is expected to be close to the anchor as soon as possible, while the whole negative sample set moves in the opposite direction.However, it ignores the structured information of confusing samples and also is not suitable for multi-modal samples.Our MMSL preserves the structured information of samples from RGB and thermal modalities as much as possible and separates all positive and negative samples from all modalities.

\section{Experiments}
To verify the effectiveness of our Multi-Modal Multi-Margin Metric Learning for RGBT Tracking(M$^5$L), we test it on two RGBT benchmark: GTOT\cite{7577747} dataset and RGBT234\cite{LI2019106977} dataset, and compare it with many state-of-the-art methods.

\subsection{Evaluation Setting}

{\flushleft \bf Datasets}.
We evaluate our M$^5$L on two common datasets, GTOT and RGBT234 datasets.  
GTOT dataset consists of 50 sequences, which are annotated with 7 attributes and coping with 7 challenges.   
RGBT234 has a total of almost 234,000 frames and 234 video sequences with different lengths.  RGBT234 is annotated with 12 attributes. 

{\flushleft \bf Evaluation metrics}.
We use two following criteria to evaluate the performance. Precision rate(PR) is the ratio of frames whose location of tracked object is within the threshold distance of  groundtruth, and success rate(SR) is the percentage of successful frames whose overlap exceed the given threshold.

\begin{figure}[htbp]
	\centering  
	\includegraphics[width=\linewidth, height=0.32\textheight]{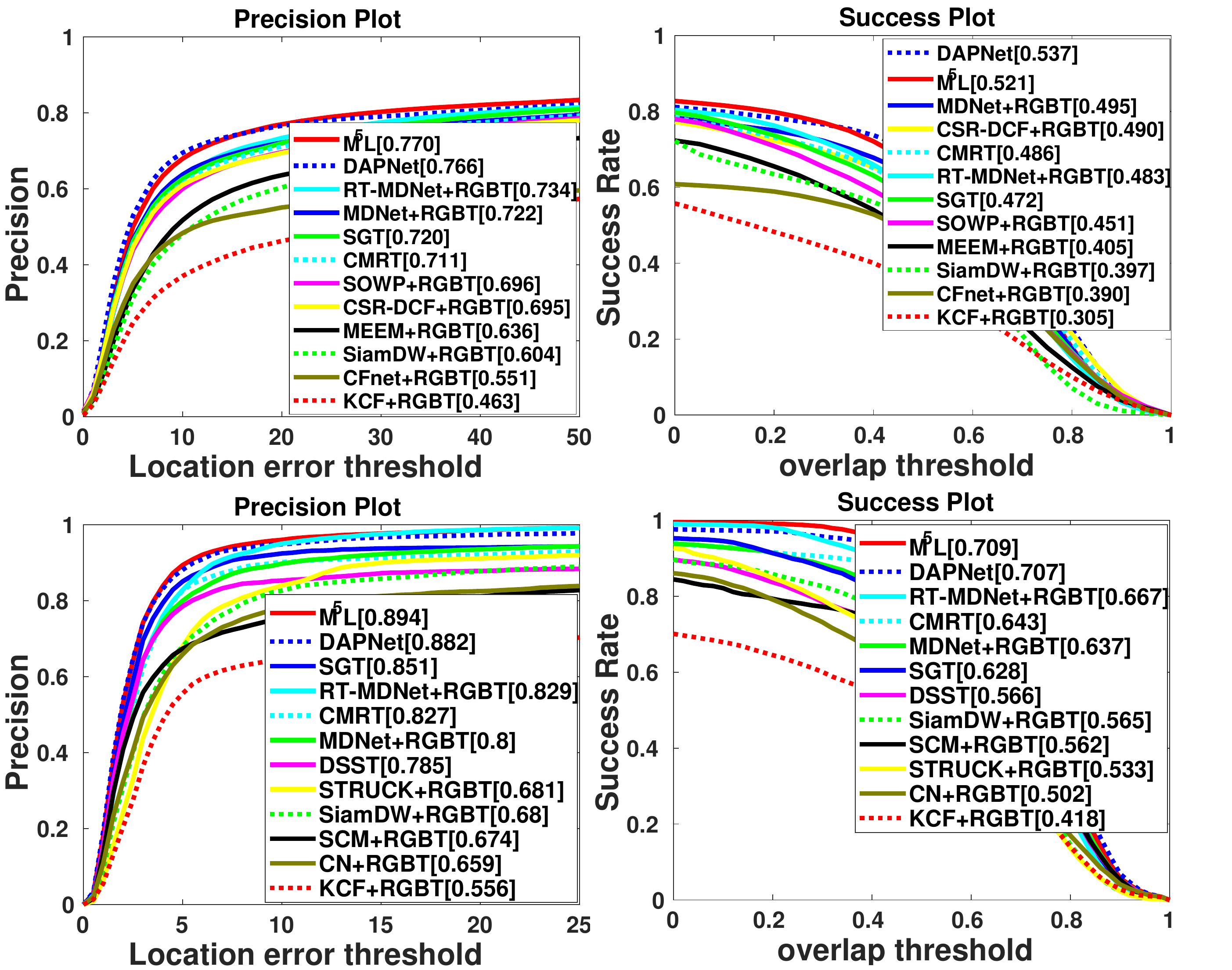}
	\caption{PR/SR scores on RGBT234 and GTOT datasets respectively. Results on RGBT234 are on the top and GTOT on the bottom.}  
	\label{eva}  
\end{figure}

\makeatletter\def\@captype{table}\makeatother
\setlength{\tabcolsep}{5pt} 
\begin{table}[!ht]\footnotesize
	\centering
	\setlength{\abovecaptionskip}{2pt}
	\setlength{\belowcaptionskip}{5pt}
	\caption{ Comparisons between M$^5$L, MDNet, DAPNet and RT-MDNet, CMRT and SiamDW. MDNet and RT-MDNet are both transformed to RGBT trackers.The best performance of the tracker is shown in bold. }\label{table5} 
	\setlength{\tabcolsep}{0.5mm}{
		\begin{tabular}{l|c|c|c c|c c|c}
			\hline
			\multirow{2}{*}{Method} &
			\multirow{2}{*}{Publication} &
			\multirow{2}{*}{Date} &
			\multicolumn{2}{c|}{RGBT234} &
			\multicolumn{2}{c|}{GTOT}&
			\multirow{2}{*}{Speed} \\
			\cline{4-7}
			& & & PR & SR & PR & SR \\
			\hline
			(a) RT-MDNet & ECCV & 2018 & {0.734} & 0.483 & {0.829} & {0.667} & 15.6fps\\
			(b) DAPNet & ACMMM & 2019    & {0.766} & \bfseries{0.537} & {0.882} & {0.707} & 2.11fps \\
			(c) MDNet & CVPR & 2016    & 0.722 & 0.483 & 0.8 & 0.637 & 1.61fps\\
			(d) CMRT & ECCV & 2018    & 0.711 & {0.486} & 0.827 & 0.643 & 8.30fps \\
			(e) SiamDW & CVPR & 2019    & 0.604 & 0.397 & 0.68 & 0.565 & 95.34fps \\
			(f) M$^5$L & ----&----& \bfseries{0.770} & {0.521} & \bfseries{0.894} & \bfseries{0.709} &\bfseries{14.26fps} \\
			\hline
	\end{tabular}}
\end{table}

\begin{figure*}[htbp]
	\centering  
	\includegraphics[width=\linewidth, height=0.76\textheight]{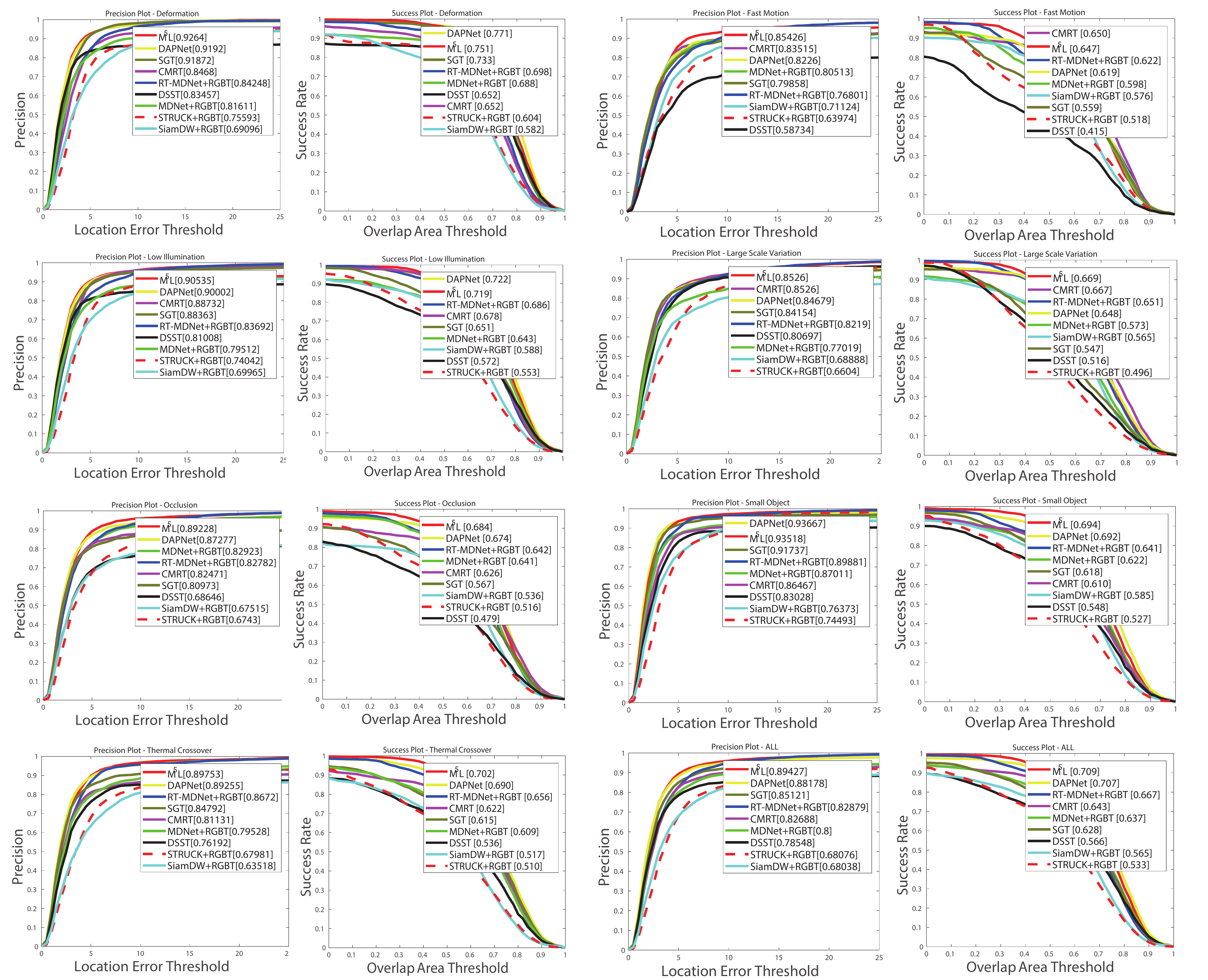}
	\caption{PR/SR evaluation results on various challenges compared with state-of-the-art methods on GTOT.}  
	\label{challenges}  
\end{figure*}

\makeatletter\def\@captype{table}\makeatother
\setlength{\tabcolsep}{5pt}
\begin{table*}[!ht]\scriptsize
	\centering
	\setlength{\abovecaptionskip}{2pt}
	\setlength{\belowcaptionskip}{5pt}
	\caption{ Comparison results of PR/SR scores(\%) of different trackers under different challenges on RGBT234.}\label{table1}
	\begin{tabular}{c|cccccccc|c}
		\hline
		{ } &{ RT-MDNet+RGBT} &{ MDNet+RGBT} ~ &{ SOWP+RGBT}~ &{ CSR-DCF+RGBT}~ &{ SiamDW+RGBT}~ &{ CFNet+RGBT}~ &{ CMRT}~ &{ DAPNet}~ &{ M$^5$L} \\
		\hline
		{ BC} & 0.725/0.455 &0.644/0.432 & 0.647/0.419 & 0.618/0.410 &0.519/0.323 &0.463/0.308 & 0.631/0.398 & 0.717/0.484 &\bfseries{0.766}/\bfseries{0.498}\\
		{CM} & 0.644/0.455 &0.640/0.454 & 0.652/0.430 & 0.611/0.445 &0.562/0.382 &0.417/0.318 & 0.629/0.447 & 0.668/0.474 &\bfseries{0.716}/\bfseries{0.500}\\ 
		{DEF} & 0.670/0.466 &0.668/0.473 & 0.650/0.460 & 0.630/0.462 &0.558/0.390 &0.523/0.367 & 0.667/0.473 & 0.717/\bfseries{0.578} &\textbf{0.727}/0.500\\
		{FM} & 0.637/0.387&0.586/0.363 & \bfseries{0.703}/\bfseries{0.435} & 0.529/0.358 &0.597/0.365 &0.454/0.299 & 0.613/0.384 & 0.670/0.443 &0.659/0.420\\
		{HO} & 0.618/0.404 &0.619/0.421 & 0.570/0.379 & 0.593/0.409 &0.520/0.337 &0.417/0.290 & 0.563/0.377 & 0.660/0.444 &\textbf{0.662}/\textbf{0.457}\\
		{LI} & 0.737/0.474 &0.670/0.455 & 0.723/0.468 & 0.691/0.474 &0.600/0.399 &0.523/0.369 & 0.742/0.498 & \bfseries{0.775}/\bfseries{0.530} &0.761/0.495\\
		{LR} & 0.760/0.483 &0.759/0.493 & 0.725/0.462 & 0.720/0.476 &0.605/0.370 &0.551/0.365 &0.687/0.420  & 0.750/\bfseries{0.510} &\textbf{0.762}/0.496\\
		{MB} & 0.612/0.429 &0.654/0.463 & 0.639/0.421 & 0.580/0.425 &0.494/0.340 &0.357/0.271 & 0.600/0.427 & 0.653/0.467 &\bfseries{0.670}/\bfseries{0.472}\\
		{NO} & 0.894/0.586 &0.862/0.611 & 0.868/0.537 & 0.826/0.600 &0.783/0.534 &0.764/0.563 & 0.895/0.616 & 0.900/\bfseries{0.644} &\textbf{0.904}/0.619\\
		{PO} & 0.780/0.517 &0.761/0.518 & 0.747/0.484 & 0.737/0.522 &0.608/0.396 &0.597/0.417 & 0.777/0.536 & 0.817/0.544 &\bfseries{0.821}/\bfseries{0.574}\\
		{SV} & 0.735/0.482 &0.735/0.505 & 0.664/0.404 & 0.707/0.499 &0.609/0.405 &0.596/0.433 & 0.710/0.493 & 0.772/0.513 &\bfseries{0.780}/\bfseries{0.542}\\
		{TC} & {\bfseries0.786}/0.513 &0.756/0.517 & 0.701/0.442 & 0.668/0.462 &0.569/0.368 &0.457/0.327 & 0.675/0.443 & 0.768/0.538 &0.781/\bfseries{0.543}\\
		\hline
		{ALL} & 0.734/0.483 &0.722/0.495 & 0.696/0.451 & 0.695/0.490 &0.604/0.397 &0.551/0.390 & 0.711/0.486 & 0.766/\textbf{0.537} &\textbf{0.770} / 0.521\\
		\hline
	\end{tabular}
\end{table*}

\subsection{Evaluation on GTOT Dataset}
{\flushleft \bf Overall performance}.
On GTOT dataset, we compare our method with existing 14 trackers,and 12 outstanding trackers are M$^5$L, RT-MDNet~\cite{Jung_2018_ECCV}+RGBT, MDNet~\cite{Nam_2016_CVPR}+RGBT, SGT~\cite{Li:2017:WSR:3123266.3123289}, KCF~\cite{6870486}+RGBT, SCM~\cite{6247882}+RGBT, DSST~\cite{danelljan2014accurate}, CN~\cite{Danelljan_2014_CVPR}+RGBT, STRUCK~\cite{7360205}+RGBT, DAPNet~\cite{zhu2019dense}, CMRT~\cite{li2018cross} and SiamDW+RGBT~\cite{zhang2019deeper} respectively, where ~\cite{Jung_2018_ECCV,Nam_2016_CVPR,Li:2017:WSR:3123266.3123289,zhu2019dense,li2018cross} are RGBT-based trackers and the remaining are RGB-based methods, we concatenate features from RGB and thermal modalities into a single vector or view the thermal modality as an extra channel for fair comparison.   
Fig.~\ref{eva} shows good performance of our approach on GTOT, our approach is better than other 11 outstanding trackers mentioned above. 
To be more specific, our method is 6.5\%/4.2\% higher than RT-MDNet+RGBT and 1.2\%/0.2\% higher than DAPNet in PR/SR.At the same time, M$^5$L has the advantage in tracking speed, which is 7 times faster than DAPNet, as shown in Table.~\ref{table5}. More details can be seen in Table.~\ref{table5}.    

{\flushleft \bf Challenges-based performance on GTOT}.
There are 7 challenges existing in GTOT, including deformation (DEF), fast motion(FM), low illumination(LI), large scale variation(LSV), occlusion(OCC), small object(SO) and thermal crossover(TC).   
The results of comparison with other state-of-the-art trackers are listed in Fig.~\ref{challenges}.   

To be more specific, our M$^5$L has beated other trackers under most challenges, including large scale variation(LSV), occlusion(OCC), small object(SO) and thermal crossover(TC), which proves the effectiveness of our proposed method.However, M$^5$L performs slightly inferior than DAPNet in terms of deformation(DEF) and low illumination(LI) as shown in Fig.~\ref{challenges}, because DAPNet aims at fusing more robust multi-modal features, which makes best use of thermal data to cope with the challenges of deformation(DEF) and low illumination(LI).On the contrary, our M$^5$L adopts a relatively simple fusion strategy which introduces modality attentions to integrate features adaptively. Besides, M$^5$L is 2.0\%/1.0\% higher than DAPNet in PR/SR in occlusion. In essence, our M$^5$L aims at distinguishing positive and negative samples, and keeping the structured information of confusing positive and negative samples, which can make the tracker locate the object easily and keep strong robustness under the challenges mentioned above. However, existing methods are weaker in discriminating the object from various disturbances compared with our M$^5$L, especially the object is extremely similar to its background or the size of object is considerably small.

\subsection{Evaluation on RGBT234 Dataset}

{\flushleft \bf Overall performance}.
To further verify the effectiveness of our approach, we evaluate our algorithm on RGBT234 dataset.   
The results of our approach compared with other 11 trackers (MDNet~\cite{Nam_2016_CVPR}+RGBT, RT-MDNet~\cite{Jung_2018_ECCV}+RGBT, CSR-DCF~\cite{7577747}+RGBT, SGT~\cite{Li:2017:WSR:3123266.3123289}, SOWP+RGBT, MEEM~\cite{Zhang2014MEEM}+RGBT, CFNet~\cite{Valmadre_2017_CVPR}+RGBT, KCF~\cite{6870486}+RGBT, DAPNet~\cite{zhu2019dense}, CMRT~\cite{li2018cross} and SiamDW+RGBT~\cite{zhang2019deeper}) are shown in Fig.~\ref{eva}, where ~\cite{Nam_2016_CVPR,Jung_2018_ECCV,Li:2017:WSR:3123266.3123289,zhu2019dense,li2018cross} are RGBT-based trackers and the remaining are RGB-based methods.   
From  Fig.~\ref{eva}, we can see that our method is 3.6\%/3.8\% higher than RT-MDNet+RGBT in PR/SR, which, to a certain extent, demonstrates the effectiveness of the proposed modules.However, our approach is 1.6\% lower than DAPNet in SR, because DAPNet uses a recursive strategy to densely aggregate features from multi-modalities. More comparisons with some most outstanding trackers are presented in Table.~\ref{table5}.

\begin{figure*}[htbp]
	\centering  
	\includegraphics[width=\linewidth,height=0.51\textheight]{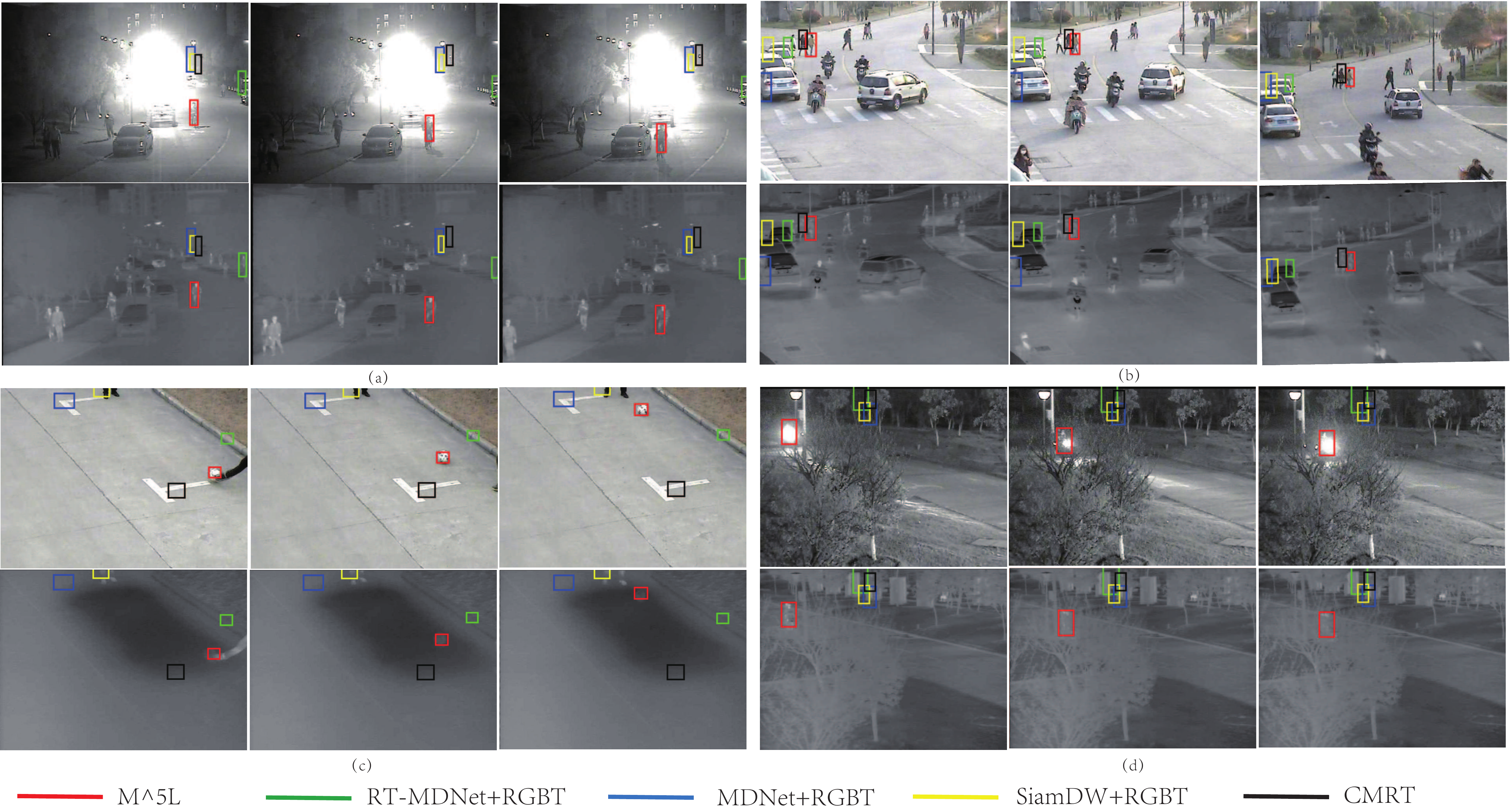}
	\caption{Comparison of our M$^5$L and other four trackers in qualitative performance on four RGBT sequences. (a),(b),(c) and (d) are corresponding to four challenges: high illumination, background cluster, fast motion and heavy occlusion respectively.}  
	\label{qualitative}  
\end{figure*}

{\flushleft \bf Challenges-based performance on RGBT234}.
There are 12 challenges in RGBT234, including background cluster (BC), camera motion (CM), deformation (DEF), fast motion (FM), heavy occlusion (HO), low illumination (LI), low resolution (LR), motion blur (MB), no occlusion (NO), partial occlusion (PO), scale variation (SV) and thermal crossover (TC).   
The results of comparison with other state-of-the-art trackers including RT-MDNet+RGBT, MDNet+RGBT, SOWP+RGBT, CSC-DCF+RGBT, MEEM+RGBT, CFNet+RGBT, DSST and KCF+RGBT, are listed in Table~\ref{table1}.  The overall performance of our M$^5$L is better than our baseline method RT-MDNet.    

In addition, our M$^5$L has achieved best performance for HO and PO, which proves that our attention-based fused features can boost the robustness of tracking in case of occlusion.   
Furthermore, BC and HO demonstrate that our feature embedding module can discriminate positive and negative samples and remain structured information of samples, which greatly helps to classify target accurately, and improves the tracking performance.   
In brief, our network performs well for background cluster, camera motion, deformation, motion blur, and many other attributes, meaning that our framework has strong robustness.  

{\flushleft \bf Qualitative performance}.
The comparison between our M$^5$L and other four trackers in term of qualitative performance has presented in Fig.~\ref{qualitative}.    
For example, In (b)and (c), our tracker performs best for partial occlusions and fast motion respectively.   The tracked object is occluded paritially by other confusing samples in (b).   By multi-margin structured loss, our M$^5$L locates the object accurately.   Bad illumination as shown in (a) and (d), our M$^5$L performs outstanding, demonstrating fused feature based on modality weight is more robust.
In (b) and (c), the human is disturbed by surrounding pedestrians so that the general trackers could not discriminate which one is the target, and likewise, when a football moves fast, football is similar to white crossings, which also disturbs the tracker. However, our M$^5$L aims at distinguishing positive and negative samples and remaining the structured information in confusing positive and negative samples, which enables tracker locate the object.

\subsection{Analysis of Our Approach}

{\flushleft \bf Mining confusing samples}.
As demonstrated in Section~\textbf{Feature embedding module}, the Multi-modal Multi-margin Structured Loss(MMSL) mines the samples from a certain interval decided by $\alpha,\beta,m$.   
Specifically, we first identify the confusing samples as shown in Eq.~(\ref{s4}) and pull confusing positive samples into the interval [$\alpha-\beta$, $\alpha$], referred in Eq.~(\ref{s8}) and Eq.~(\ref{s10}).   
Simultaneously, the optimization can pull the confusing negative samples into [$\alpha+m$,$\alpha+m+\beta$], referred in Eq.~(\ref{s9}) and Eq.~(\ref{s11}).   
Hence, we conduct some experiments on RGBT234 dataset to analyze the effects of pre-defined parameters.

\makeatletter\def\@captype{table}\makeatother
\setlength{\tabcolsep}{5pt}
\begin{table}[!ht]\normalsize
	\centering
	\setlength{\abovecaptionskip}{2pt}
	\setlength{\belowcaptionskip}{5pt} 
	\caption{ Influence of $m$ between positive and negative samples in the Multi-modal Multi-margin Structured Loss.   PR/SR results(\%) are listed in the table with $\alpha$=1.6, $\beta$=0.1.}\label{table2}
	\begin{tabular}{ccc}
		\hline
		{$\alpha$=1.6, $\beta$=0.1} &{ Precision Rate} &{ Success Rate} \\
		\hline
		{$m$ = 0} & 0.745 &0.506 \\
		{$m$ = 0.2} & \bfseries0.770 &\bfseries0.521 \\
		{$m$ = 0.4} & 0.744 &0.496 \\
		\hline
	\end{tabular}
\end{table}

\makeatletter\def\@captype{table}\makeatother
\setlength{\tabcolsep}{5pt}
\begin{table}[!ht]\normalsize
	\centering
	\setlength{\abovecaptionskip}{2pt}
	\setlength{\belowcaptionskip}{5pt} 
	\caption{ Influence of $\beta$ which controls the size of fixed intervals.PR/SR results(\%) are showed in the table with $\alpha$=1.6, $m$=0.2.}\label{table3}
	\begin{tabular}{ccc}
		\hline
		{$\alpha$=1.6, $m$=0.2} &{ Precision Rate} &{ Success Rate} \\
		\hline
		{$\beta$ = 0} & 0.753 &0.514 \\
		{$\beta$ = 0.1} & \bfseries0.770 &\bfseries0.521 \\
		{$\beta$ = 0.2} & 0.753 &0.504 \\
		{$\beta$ = 0.3} & 0.741 &0.492 \\
		\hline
	\end{tabular}
\end{table}

{\flushleft \bf Influence of  $m$}.
First, we fix $\alpha$=1.6 and $\beta$=0.1.
From Table~\ref{table2}, it is easily seen that when $m$=0.2, the Multi-modal Multi-margin Structured Loss performs much better than other margins, i.e., 3\% higher in PR/SR.   
Therefore, it is important for Multi-modal Multi-margin Structured Loss to select a proper margin.   
More details are presented in Table~\ref{table2}.

{\flushleft \bf Influence of $\beta$}.
To study the effect of $\beta$, we fix $\alpha$=1.6 and $m=0.2$.   
From Table~\ref{table3}, we verify our idea that we optimize the confusing positive and negative samples into the fixed intervals and can keep the structured information of all samples to improve tracking performance.

\begin{figure}[htbp]
	\centering  
	\includegraphics[width=0.35\textheight,height=0.32\textheight]{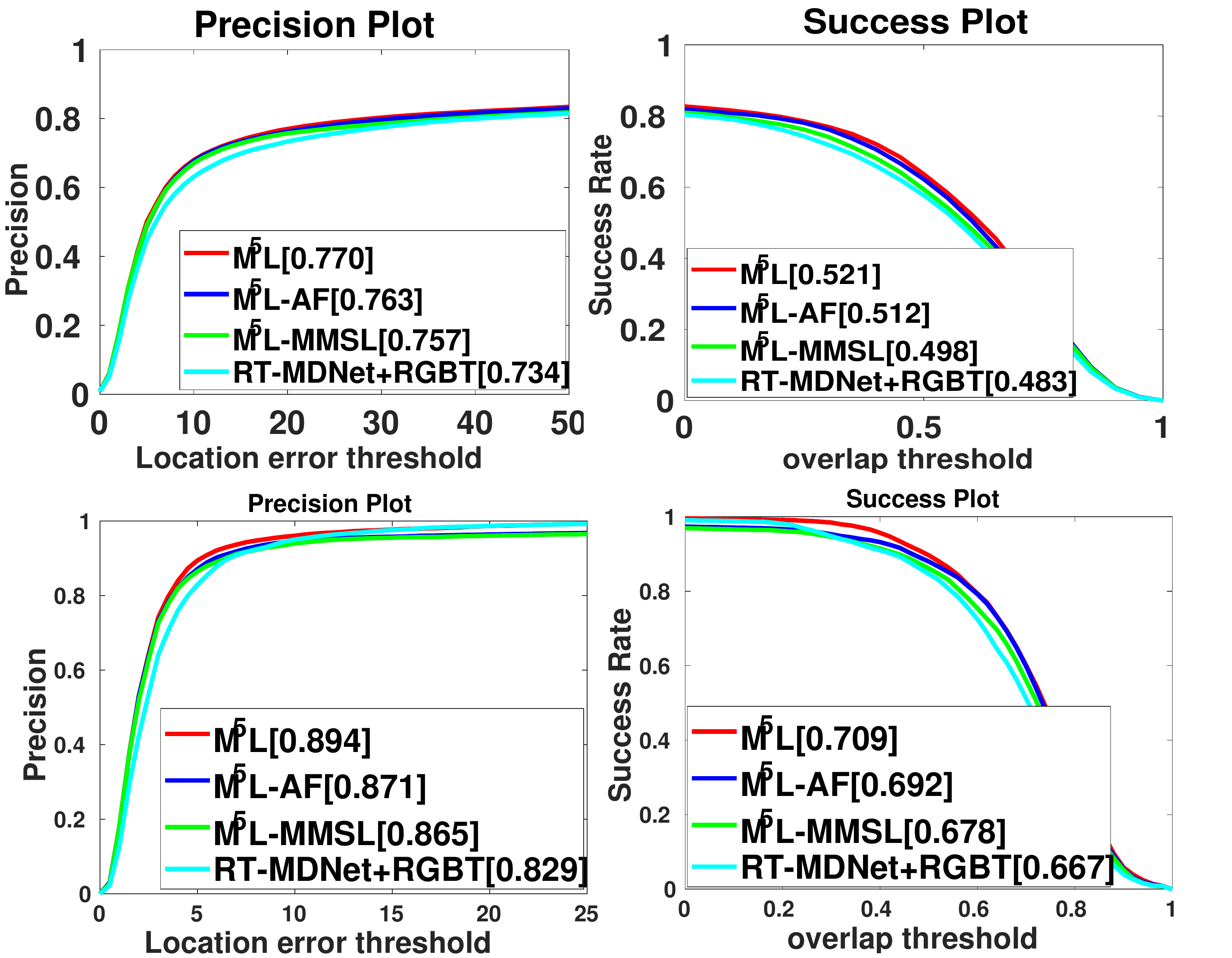}
	\caption{Performance evaluation without attention-based fusion(AF)/ Multi-modal Multi-margin Structured Loss(MMSL) on datasets RGBT234 and GTOT respectively.Ablation study on RGBT234 is on the top and GTOT on the bottom. }  
	\label{ab}  
\end{figure}

{\flushleft \bf Ablation Study}.
To justify main components of the proposed network, we evaluate attention-based fusion module and feature embedding module on  GTOT and RGBT234 datasets.   
We successively remove attention-based fusion module (M$^5$L-AF) and feature embedding module (M$^5$L-MMSL).   
Fig.~\ref{ab} presents the results on RGBT234 and GTOT respectively.   
The superior performance of M$^5$L versus M$^5$L-AF, M$^5$L-MMSL and our baseline RT-MDNet verifies the effectiveness of two components to predict modality attentions and effectively separate positive and negative samples.
Table.~\ref{table4} illustrates $L_{cross}$ and $L_{rgb},L_t$ are all beneficial for RGBT tracking.

\makeatletter\def\@captype{table}\makeatother
\setlength{\tabcolsep}{5pt}
\begin{table}[!ht]\normalsize
	\centering
	\setlength{\abovecaptionskip}{2pt}
	\setlength{\belowcaptionskip}{5pt}
	\caption{ Results evaluated on two RGBT datasets of RGBT234 and GTOT.We make comparison between M$^5$L, M$^5$L w/o $L_{cross}$, M$^5$L w/o $L_{rgb},L_{t}$ and the baseline.}\label{table4}
	\begin{tabular}{l|c c|c c}
		\hline
		\multirow{2}{*}{Method} &
		\multicolumn{2}{c|}{RGBT234} &
		\multicolumn{2}{c}{GTOT} \\
		\cline{2-5}
		& PR & SR & PR & SR \\
		\hline
		(a) $Baseline$ & 0.734 & 0.483 & 0.829 & 0.667 \\
		(b) $M^5L$    w/o $L_{cross}$ & 0.751 & 0.506 & 0.863 & 0.691 \\
		(c) $M^5L$    w/o $L_{rgb},L_t$ & 0.745 & 0.493 & 0.877 & 0.683 \\
		(d) $M^5L$ & \bfseries0.770 & \bfseries0.521 & \bfseries0.894 & \bfseries0.709 \\
		\hline
	\end{tabular} 
\end{table}

{\flushleft \bf Runtime analysis}.
Ultimately, we analyze the runtime of our M$^5$L versus the baseline RT-MDNet and RT-MDNet+RGBT. 
Our implementation is on PyTorch0.4.0, python2.7 with NVIDIA GeForce GTX 1080Ti GPU and 4.2GHz intel Core i7-8700k.  
The runtime of RT-MDNet, RT-MDNet+RGBT and M$^5$L on RGBT234 is 25fps, 15fps and 14fps respectively.   
However, their performances are far behind our M$^5$L, meanwhile, M$^5$L is much faster than MDNet (14 times faster than MDNet 1fps).   

\section{Conclusion}
In this paper, we have proposed a novel deep metric learning framework for RGBT tracking.  
We adopt an attention-based fusion strategy to adaptively aggregate features of different modalities, 
and also design an effective feature embedding module with a new deep metric loss called Multi-modal Multi-margin Structured Loss to distinguish confusing samples and remain structured information of all samples in each modality, and cross-constraint is further adopted to separate positive and negative samples from different modalities, which could boost the tracking performance.   
Extensive experiments have suggested that our tracker achieves the superior tracking performance against other state-of-the-art trackers.

\ifCLASSOPTIONcaptionsoff
  \newpage
\fi

\bibliographystyle{IEEEtran}
\bibliography{mybib}

\end{document}